\documentclass[10pt,twocolumn,letterpaper]{article}

\usepackage{cvpr}
\usepackage{times}
\usepackage{epsfig}
\usepackage{graphicx}
\usepackage{amsmath}
\usepackage{amssymb}
\usepackage{float}
\usepackage{subcaption}
\usepackage{multicol}

\usepackage[pagebackref=true,breaklinks=true,letterpaper=true,colorlinks,bookmarks=false]{hyperref}

\cvprfinalcopy 

\def\cvprPaperID{5300} 

\ifcvprfinal\fi
\begin{document}

\title{Joint Pose and Shape Estimation of Vehicles from LiDAR Data}

\author{Hunter Goforth\textsuperscript{1} \quad Xiaoyan Hu\textsuperscript{2} \quad Michael Happold\textsuperscript{1} \quad Simon Lucey\textsuperscript{1,3} \\\\ \textsuperscript{1}Argo AI \quad \textsuperscript{2}Microsoft \quad \textsuperscript{3}Carnegie Mellon University \\\\
{\small \texttt{\{hgoforth, mhappold\}@argo.ai \quad xiaoyan.hu@microsoft.com \quad slucey@cs.cmu.edu}}
}


\twocolumn[{%
\renewcommand\twocolumn[1][]{#1}%
\maketitle
\begin{center}
    \centering
    \includegraphics[width=\textwidth]{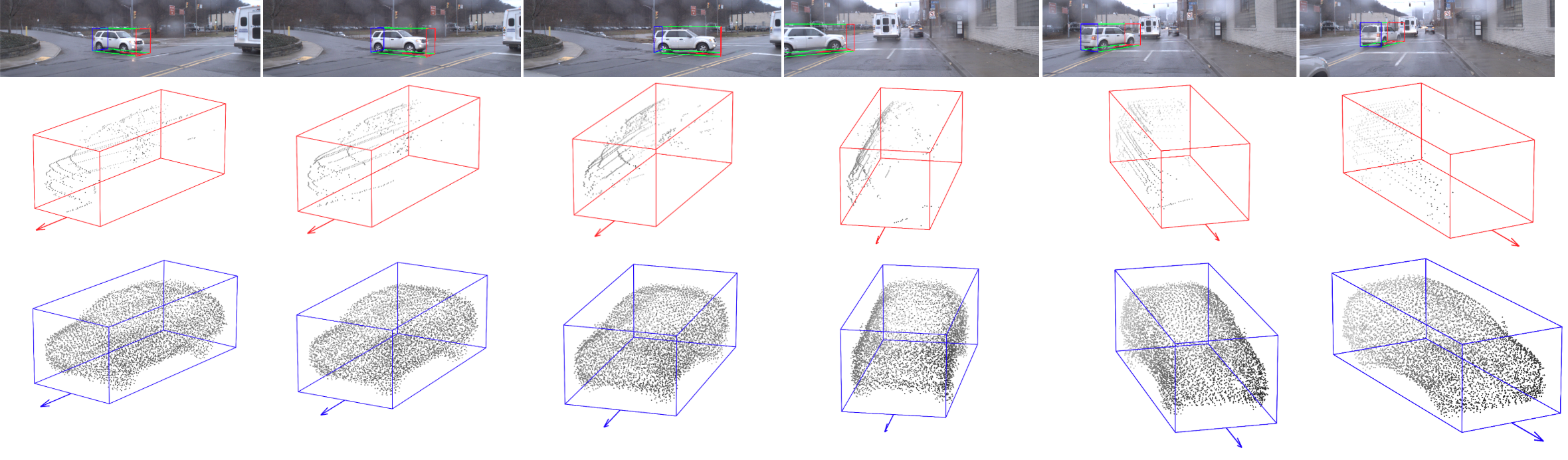}
    \captionof{figure}{We propose a method which produces accurate pose and shape estimation of vehicles (bottom row). RGB images are for visual reference, our method relies only upon the unaligned, sparse LiDAR input shown in the second row (ground truth pose is shown in red). Example data from Argoverse~\cite{chang2019argoverse}.}
    \label{fig:teaser}
\end{center}%
}]

\begin{abstract}
We address the problem of estimating the pose and shape of vehicles from LiDAR scans, a common problem faced by the autonomous vehicle community. Recent work has tended to address pose and shape estimation separately in isolation, despite the inherent connection between the two. We investigate a method of jointly estimating shape and pose where a single encoding is learned from which shape and pose may be decoded in an efficient yet effective manner. We additionally introduce a novel joint pose and shape loss, and show that this joint training method produces better results than independently-trained pose and shape estimators. We evaluate our method on both synthetic data and real-world data, and show superior performance against a state-of-the-art baseline.
\end{abstract}

\section{Introduction}

At the core of all autonomous vehicle systems today is a strong perception backbone, which is capable of identifying relevant agents in the environment. A key requirement of the perception system is to identify other vehicles sharing the road, and reason about the pose and shape of these vehicles. Autonomous vehicles are equipped with many sensing modalities, including LiDAR, which is the modality on which we focus.

Often, the first step in LiDAR processing is to perform some form of clustering or segmentation, to isolate parts of the raw point cloud which belong to individual objects. This can be done using only the LiDAR point cloud~\cite{wu2018squeezeseg,milioto2019rangenet++}, but can also be done by using masks back-projected from RGB~\cite{qi2018frustum}, or with other methods~\cite{wang2018pointseg,biasutti2019riu}. Following the segmentation step, we are left with individual segments belonging to cars, pedestrians, and other classes. Examples of LiDAR segments of cars are shown in the middle row of Fig.~\ref{fig:teaser}.

For segments belonging to vehicles, the next step is to infer both the pose and shape of the vehicle. In prior work such as Frustum PointNet~\cite{qi2018frustum}, this is done by directly regressing to parameters of an amodal bounding box, \textit{lacking an explicit estimation of shape}. A primary contribution of our work is to present the first pipeline for explicitly determining pose and shape in a joint manner from LiDAR data. The benefit of such an approach is twofold: (1) our modularized approach provides clarity and explanation for why a certain amodal bounding box might be predicted, by explicitly representing shape and (2) the estimated shape model is available for use in downstream modules such as tracking.

Recent literature deals extensively with finding both pose and shape from partial observations using the PointNet~\cite{qi2017pointnet} architecture. Point Completion Network~\cite{yuan2018pcn} deals with finding shape from partial observation, but makes the assumption that \textit{pose is known beforehand}, and all observations are in a canonical orientation. PointNet-based registration methods~\cite{aoki2019pointnetlk,gross2019alignnet} deal with finding pose of partial observations, however these works assume the existence of a complete template to which they can register the partial observation. IT-Net~\cite{yuan2018iterative} addresses the problem of registration without a template, where a partial observation may be aligned to a canonical orientation. In this work, we build upon these past efforts, building a PointNet-based pipeline which generates both a completed point cloud representing the vehicle shape, while also finding the pose of the vehicle \textit{in a joint manner which shares information between these two tasks.}

Training a supervised model to predict complete shape from partial observations requires a large dataset of partial and complete observations. To meet this requirement, we constructed a large dataset using high-resolution CAD models of various vehicle types. We sampled points from these CAD models using a simulated LiDAR sensor, and trained networks to produce a complete point cloud representing a uniform surface sampling from the exterior of the vehicles. We use this dataset for training and validation of baseline models as well as our proposed models. In addition to evaluating on this simulated dataset, we also provide inference results on real-world data from the Argoverse dataset~\cite{chang2019argoverse}. We select examples from Argoverse for which we have access to the ground truth CAD model, so we are able to quantify the completion quality of real-world data.

\textbf{Contributions.} In this work, we develop a novel network architecture for jointly estimating shape and pose of vehicles from partial LiDAR observations. We also develop a novel joint loss combining pose and shape estimation, and show that this joint training regime improves performance over independently-trained shape and pose modules. We show that our architecture achieves state-of-the-art results on realistic synthetic data as well as on real-world data, comparing against a previous state-of-the-art baseline. In addition to being more accurate, our network architecture uses many less parameters and therefore provides efficient inference, addressing a key concern of the autonomous vehicle community.

\begin{figure*}
    \centering
    \includegraphics[width=\textwidth]{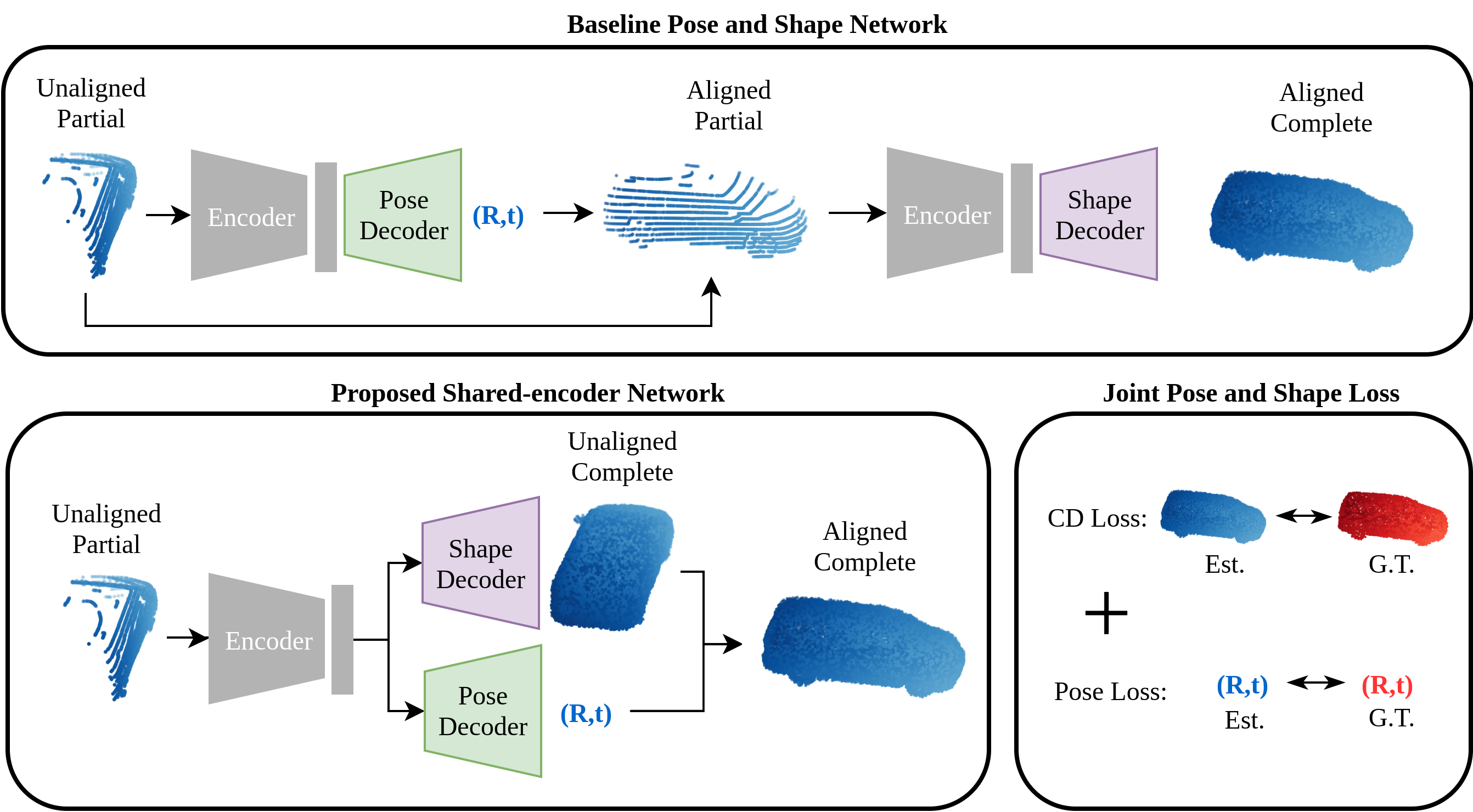}
    \caption{\textbf{Baseline and proposed network architectures.} The baseline for predicting pose and shape first estimates the pose of a unaligned partial input, then applies this pose to the partial input before estimating the shape. However, in this configuration, the shape estimation is subject to errors in the pose estimation, and the partial input is redundantly encoded twice. In the proposed shared-encoder network, a single encoding is used to predict both shape and pose, reducing redundancy and providing stable pose and shape estimation. Training the shared-encoder network is done by first training the encoder for shape completion, before training the pose decoder on codes produced by a frozen encoder. We show that performance can be further improved through joint-training using a combined pose and shape loss. We adopt the encoder and shape decoder design of PCN~\cite{yuan2018pcn}, while the pose decoder network constitutes an MLP of size (1024,512,512,3) which predicts a heading $\theta$ and translation $(x,y)$.}
    \label{fig:network_arch}
\end{figure*} 

\section{Related Work}

\textbf{Shape estimation.} Much work has been done on the problem of estimating full 3D shape from a partial observation of a depth sensor. For applications where only local hole-filling in the partial observation is needed, many interpolation-based methods exist~\cite{thanh2016field,sorkine2004least,zhao2007robust}, however these methods require a certain level of completeness in the partial scan. For more extreme cases of incompleteness, methods have been developed to utilize symmetry heuristics~\cite{thrun2005shape,mitra2006partial,sipiran2014approximate,tatarchenko2018tangent}, or to utilize a dataset of existing object or part models for matching~\cite{li2015database,pauly2005example,shao2012interactive,kim2013learning,martinovic2013bayesian,sung2015data}. However, these existing methods often fall short in real-world occlusion scenarios where appropriate symmetric cues may be unavailable, or a model does not exist in the dataset which is a best-fit for the observation.

Recent work in shape completion has turned toward data-driven neural network solutions. These parameterized models have been shown to be able to learn shape cues (such as symmetry) while also being robust to occlusion and novel objects. These approaches differ in the volumetric representation used: \cite{wu2018learning} proposes a voxel-based method which can be memory-intensive due to 3D convolution, while~\cite{litany2018deformable} propose a mesh-based method focused on human shape which may not be easily extensible for broader application. We advocate for the approach used in Point Completion Network (PCN)~\cite{yuan2018pcn}, which is a point-based method proven to be lightweight, fast, and broadly applicable.

\textbf{Pose estimation.} Estimating the 6D pose of a known rigid object has been widely explored in literature for many years. Recent approaches commonly utilize RGB only~\cite{tremblay2018deep,wu2018real,xiang2017posecnn}, or a combination of RGB and depth~\cite{wang2019densefusion,michel2017global,wong2017segicp}. However, these approaches inherently assume a known 3D shape against which partial measurements can be matched and aligned.

More difficult is the problem of estimating the pose of a unknown object of a known class (such as ``car''), although recently the problem has received more interest with the introduction of autonomous vehicles. The problem is ill-posed and requires knowledge of a ``canonical'' orientation of a class. This has been explored through the use of Spatial Transformer Networks~\cite{jaderberg2015spatial}, which learn to align inputs to a canonical axis to enhance accuracy of downstream tasks, such as classification. Canonical alignment was further explored with Iterative Transformer Network~\cite{yuan2018iterative}, a dedicated network for determining pose of objects of a known class.

\textbf{Joint shape and pose.} Several recent works attempt to solve the problem of jointly finding shape and pose, in the context of autonomous driving. \cite{ding2018vehicle} proposes a monocular-based method which generates a wire-frame model only and relies upon wide baseline matching. The methods proposed in~\cite{coenen2019probabilistic,kundu20183d,manhardt2019roi} attempt to fit morphable shape models to the observed vehicle, making them reliant upon having similar vehicle representations in a database and limiting their usage on novel vehicles (semi-trucks, cement trucks, etc.). All prior methods have focused on image data, and we present the first to focus on LiDAR input.

\section{Pose and Shape Networks}

In this section we introduce the network architectures which we will further investigate in this work, which are depicted in Fig.~\ref{fig:network_arch}. First, we describe the state-of-the-art baseline for generating both a pose and completion output from a partial LiDAR observation. Next, we introduce a novel architecture for generating both pose and completion output in a joint manner, which we call the \textbf{Shared-Encoder} network. Finally, we describe a novel training process which further enhances performance of our joint architecture. We denote this final network as \textbf{Jointly-Trained Shared-Encoder} network.

\subsection{Baseline}

Point Completion Network~\cite{yuan2018pcn} introduced an encoder-decoder architecture for point completion which has achieved state-of-the-art performance. In the original work, the authors experiment only with partial observations which are in a canonical orientation. In effect, this assumes that the pose of the observation is previously known. Thus, to build our baseline we first construct a pose estimator consisting of an encoder borrowed from PCN~\cite{yuan2018pcn} and a decoder which consists of an MLP of shape (1024,512,512,3), predicting heading $\theta$ and translation $(x,y)$. A similar method of pose estimation of a partial point cloud was also used in Frustum PointNets~\cite{qi2018frustum}. We apply this predicted pose to the partial observation, bringing the observation into a canonical coordinate system. We then input this canonically-aligned partial observation into an encoder-decoder completion network which has been trained to perform completion of canonically-aligned partial inputs.

\textbf{Weaknesses}. There are two aspects of this baseline which affect performance in the real world. The first issue, is that any error in the output of the pose estimation network will propagate through to the completion network (which has been trained only on canonically-aligned input), leading to poor completion performance. The second issue is that the partial point cloud is encoded twice (once during pose estimation and once during completion), which we consider to be redundancy in computation which may be avoided with an alternative network design. It is with these two weaknesses in mind that we develop the proposed network architecture.

\subsection{Shared-encoder Network}

Our insight is to use a shared-encoder network for estimating pose and shape. Following the basic encoder-decoder strategy for completion, we first observe that \textbf{such an architecture can also be trained on partial observations which are not canonically-aligned}. In addition, when training an encoder-decoder network for completion, we hypothesize that \textbf{the intermediate encoder result will be a useful feature from which the pose of the partial input can also be decoded}, avoiding the redundancy of encoding the partial input twice. Through these two observations, we directly address the weaknesses of the state-of-the-art baseline, and develop the shared-encoder network architecture for estimation of pose and shape shown in Fig.~\ref{fig:network_arch}.

\subsection{Training Strategy} \label{training_strategy}

\textbf{Shape training}. We begin training the shared-encoder network by freezing the pose decoder weights, and training only the encoder and shape decoder on partial inputs which are unaligned. We use Chamfer Distance (further described in Sec.~\ref{loss_functions}) as a loss function. Thus, the encoder learns to abstract the unaligned partial input into a fixed-length code which captures the vehicle shape in such a way as the complete shape can be recovered in the same (unknown) pose as the input.

\textbf{Pose training}. After training the encoder and completion decoder, we hypothesize that the codes generated from the encoder must also capture information about pose which may also be decoded. To test this hypothesis, we freeze the encoder and train the pose decoder to estimate poses of partial inputs, using ``frozen'' codes. As will be shown, the performance of the pose estimator trained in this manner can match or outperform the pose estimation of the baseline network. We call this network the \textbf{shared-encoder} model.

\textbf{Joint training}. Finally, we unfreeze all parts of the network (encoder, completion decoder, and pose decoder) and train all parts together using a joint loss function described in Sec.~\ref{loss_functions}. We find that the jointly-trained network outperforms both the baseline and standard shared-encoder networks in accuracy of both completion and pose estimation. We call this network the \textbf{jointly-trained shared-encoder} model.

\subsection{Loss Functions} \label{loss_functions}

\textbf{Shape loss}. For training completion networks, we use the Chamfer Distance (CD) between the estimated and ground truth point completions,

\begin{equation}
    \begin{split}
        L_{CD}(X, \widetilde{X}) = &\frac{1}{\lvert X \rvert} \sum_{x \in X} \min_{y \in \widetilde{X}} \lvert \lvert x - y \rvert \rvert_2 \\
        &+ \frac{1}{\lvert \widetilde{X} \rvert} \sum_{y \in \widetilde{X}} \min_{x \in X} \lvert \lvert y - x \rvert \rvert_2,
    \end{split}
\end{equation}

where $X$ is the ground truth completed point cloud and $\widetilde{X}$ is the estimated completion. We choose CD instead of Earth Mover Distance (EMD), because CD penalizes global structure instead of point density~\cite{yuan2018pcn}, and does not require one-to-one correspondence between points. In the autonomous vehicle use-case, it is most important that the global structure is correct in the shape estimate.

\textbf{Pose loss}. For training pose networks, we use a loss function originally proposed in~\cite{xiang2017posecnn}, and subsequently used in~\cite{yuan2018iterative}:

\begin{equation} \label{pose_loss}
    L_P((R,\textbf{t}),(\widetilde{R},\widetilde{\textbf{t}})) = \frac{1}{\lvert X \rvert} \sum_{x \in X} \lvert \lvert (Rx+\textbf{t}) - (\widetilde{R}x+\widetilde{\textbf{t}}) \rvert \rvert^2_2, 
\end{equation}

where $R \in SO(3), \textbf{t} \in \mathbb{R}^3$ are the ground truth pose and $\widetilde{R} \in SO(3), \widetilde{\textbf{t}} \in \mathbb{R}^3$ the estimated pose, and $X$ the ground truth complete point cloud.  In practice, for vehicle pose estimation, two degrees of freedom in rotation (pitch and roll) are often assumed to be zero, and one degree of freedom in translation (height) is assumed to be known. Thus, in subsequent results we show pose estimation results using yaw $\theta$ and translation $(x,y)$. This loss function has the benefit that it accounts for rotation and translation equally without complicated weighting in the loss function.

\textbf{Joint Loss}. For joint training, we must use a loss which combines both completion and pose loss. To avoid a parameter search for optimal multi-task weights, we adopt the strategy of learned uncertainty-based weighting proposed in~\cite{kendall2018multi}. Thus, we formulate the loss as

\begin{equation} \label{multitask_loss}
    \begin{split}
        L_J&(X,\widetilde{X},(R,\textbf{t}),(\widetilde{R},\widetilde{\textbf{t}})) = \\ &\frac{1}{2\sigma_{CD}^2} L_{CD}(X, \widetilde{X})
        + \frac{1}{2\sigma_{P}^2} L_P((R,\textbf{t}),(\widetilde{R},\widetilde{\textbf{t}})) \\
        &+ \log{\sigma_{CD}\sigma_{P}},
    \end{split}
\end{equation}

where $\sigma_{CD}$ and $\sigma_{P}$ are learned parameters representing the uncertainty of the pose and completion predictions. A larger magnitude for these terms represents a greater uncertainty in the prediction, and thus a lowered weight for the particular loss term. The $\log$ term prevents the uncertainties from becoming too large.

\begin{table}
\begin{center}
    \begin{tabular}{||c c||} 
        \hline
        \textbf{Type} & \textbf{Count} \\ [0.5ex] 
        \hline\hline
        Sedan & 59 \\
        \hline
        Large Truck & 52 \\
        \hline
        Coupe & 43 \\
        \hline
        SUV & 39 \\
        \hline
        Van & 20 \\
        \hline
        Bus & 19 \\
        \hline
        Truck & 13 \\
        \hline
        Misc. & 23  \\
        \hline
        \textbf{Total} & \textbf{278} \\
        \hline
    \end{tabular}
    \caption{\textbf{Vehicle types in our synthetic dataset.} Miscellaneous category includes garbage trucks, ambulance, fire truck, cement mixer, road cleaner, snow plow, tow truck, excavator, and RV.}
    \label{tab:hum3d_table}
\end{center}
\end{table}

\section{Synthetic Dataset} \label{simulated_dataset}

For training supervised networks which estimate complete point clouds from partial observations, it is required to have a dataset consisting of many pairs of partial observations and complete ground truth. Obtaining complete ground truth for real-world data is time-consuming, and thus we opt for training networks on accurately-simulated depth data. This section explains details of the simulated vehicle dataset used.

\subsection{CAD Models}

We collected 278 vehicle CAD models from \href{https://hum3d.com/}{Hum3D}, a vendor for high-resolution models. Most vehicles were chosen from the top-selling cars, trucks, and SUVs from the last decade, along with other categories such as semi-trailers and motorcycles. The statistics by category for our CAD models are shown in Table~\ref{tab:hum3d_table}.

\subsection{Synthetic Partial Data}

To generate partial observations in simulation, we use a Velodyne HDL-32E sensor model which is implemented by the Blensor software package~\cite{gschwandtner2011blensor}, inside the Blender rendering environment~\cite{blender}. We sample points from the surface of the CAD models using the HDL-32E simulated sensor. We place the sensor at a height of 2 meters (the approximate height of a sensor placed on top of a vehicle), and simulate vehicle poses between 5 and 35 meters distance from the sensor, with random headings uniformly sampled between 0 and 360 degrees. We generate 128 different views for each of the 278 CAD models, creating a dataset of 35,584 pairs. We reserve 15 randomly-selected vehicle models (1,920 pairs) for validation, and leave the rest for training.

\subsection{Synthetic Complete Data}

To generate complete point clouds for the CAD models, we used the \texttt{mesh2pcd} function in the PCL library~\cite{rusu2011pcl}, which generates a uniform surface sampling of the CAD model using ray tracing to exterior surfaces.

\section{Synthetic Experiment} \label{sim_exp}

In this experiment, we train and validate the baseline network, shared-encoder network, and jointly-trained shared-encoder network on the synthetic dataset. We show that our shared-encoder network outperforms the baseline architecture in completion quality as well as pose estimation, despite having fewer parameters. We additionally show that the jointly-trained shared-encoder network can further improve accuracy of both completion and pose estimation.

\subsection{Training Details}

All training uses the Adam optimizer with a learning rate of $1e^{-4}$ and batch size 32. All completion networks output a point cloud containing 16,384 points, and all pose predictors output a heading $\theta$ and translation $(x,y)$.

\textbf{Baseline}. We train the completion encoder-decoder of the baseline network to perform completion on canonically-aligned data from the synthetic dataset, saving the network which achieves the lowest validation error. We train the pose encoder-decoder of the baseline network also using the synthetic dataset, which canonically aligns an input partial point cloud. We again save the pose encoder-decoder with the lowest validation error. The baseline network consists of the pose estimator followed by the shape estimator, each of which have been trained independently of one another.

\textbf{Shared-encoder}. We train the shared-encoder network as described in Sec.~\ref{training_strategy} using the synthetic dataset. We first train the encoder and completion decoder to perform completion of partial inputs which are not canonically aligned. Then, we freeze the encoder and train the pose decoder using the codes produced by the frozen encoder. We save the encoder and shape-decoder weights which produce the lowest Chamfer-Distance validation error, and save the pose decoder weights which produce the lowest validation error using the pose loss in (\ref{pose_loss}).

\textbf{Jointly-trained Shared-encoder}. We jointly-train the best performing shared-encoder network using the multitask loss in (\ref{multitask_loss}), saving the model with lowest validation loss.

\begin{figure*}
    \centering
    \begin{subfigure}[c]{0.32\textwidth}
        \centering
        \includegraphics{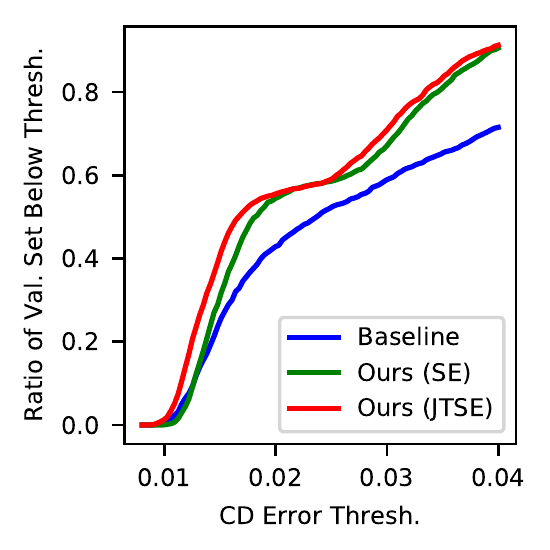}
    \end{subfigure}
    \begin{subfigure}[c]{0.32\textwidth}
        \centering
        \includegraphics{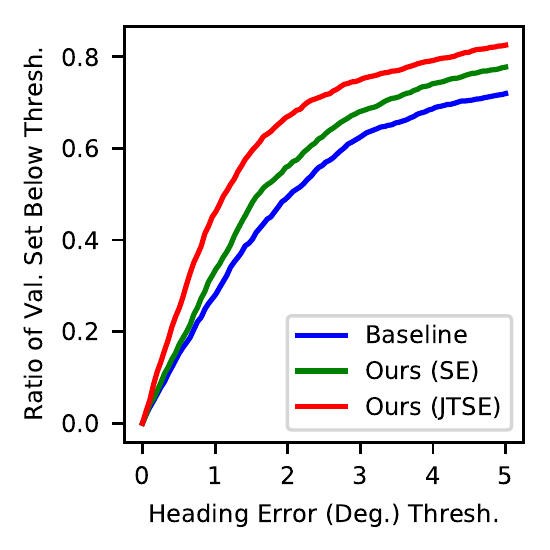}
    \end{subfigure}
    \begin{subfigure}[c]{0.32\textwidth}
        \centering
        \includegraphics{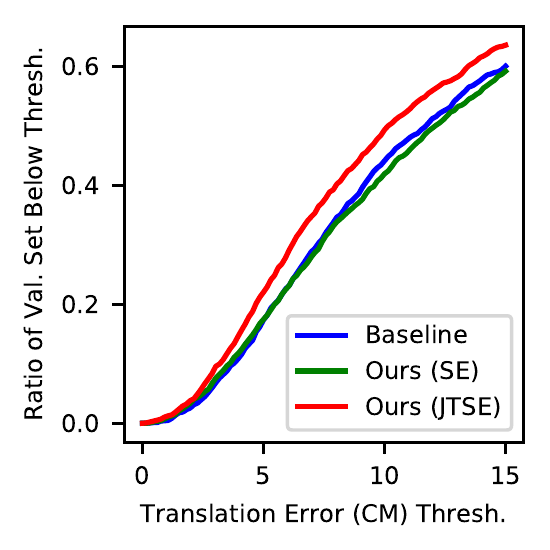}
    \end{subfigure}

    \caption{\textbf{Completion and pose accuracy on synthetic dataset.} Our shared-encoder (SE) and joint-trained shared-encoder (JTSE) networks outperform the state-of-the-art baseline in completion quality (measured as Chamfer Distance) as well as pose estimation (heading angle and translation) on the validation set from our synthetic dataset. For each metric, JTSE gives the highest ratio of the validation set under a given error threshold, followed by SE. (Left) Our shared-encoder architecture enjoys robust completion accuracy as it is not subject to errors in the pose estimation. (Center) Heading error can be more accurately estimated using the shared-encoder architecture, and further improved through joint-training. (Right) Using the shared-parameter architecture is as-good or better at estimating accurate translation compared to the baseline, despite using fewer network parameters and learning pose from pretrained encodings.\newline}
    \label{fig:hum3d_statistics}
\end{figure*}

\begin{figure*}
    \centering
    \includegraphics[width=\textwidth]{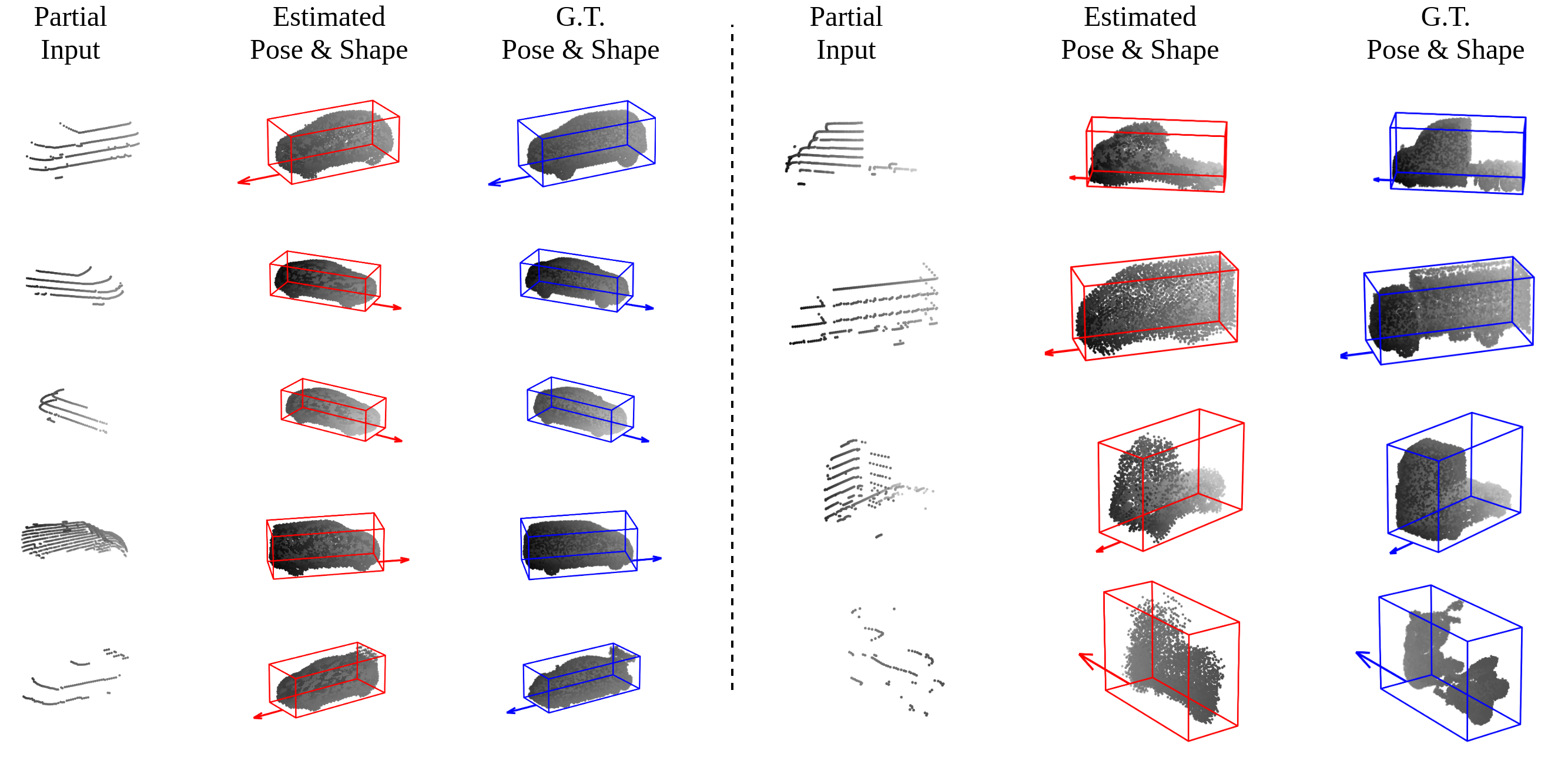}
    \caption{\textbf{Qualitative results on our synthetic dataset.} Our jointly-trained shared-encoder network produces accurate pose and shape estimation on data generated using a synthetic LiDAR sensor and vehicle CAD models. All examples are from data unseen during training. The network is able to learn to predict accurate vehicle heading from only the input partial LiDAR scan, as well as accurate shape models. In the right column, we see robustness to rarely-seen vehicle classes such as semi-trailer and motorcycle. All results are generated using a single network, although separate networks could be trained for individual vehicles clases (e.g. small cars, large trucks, motorcycles, etc.)}
    \label{fig:hum3d_qualitative}
\end{figure*}

\begin{figure*}
    \centering
    \includegraphics{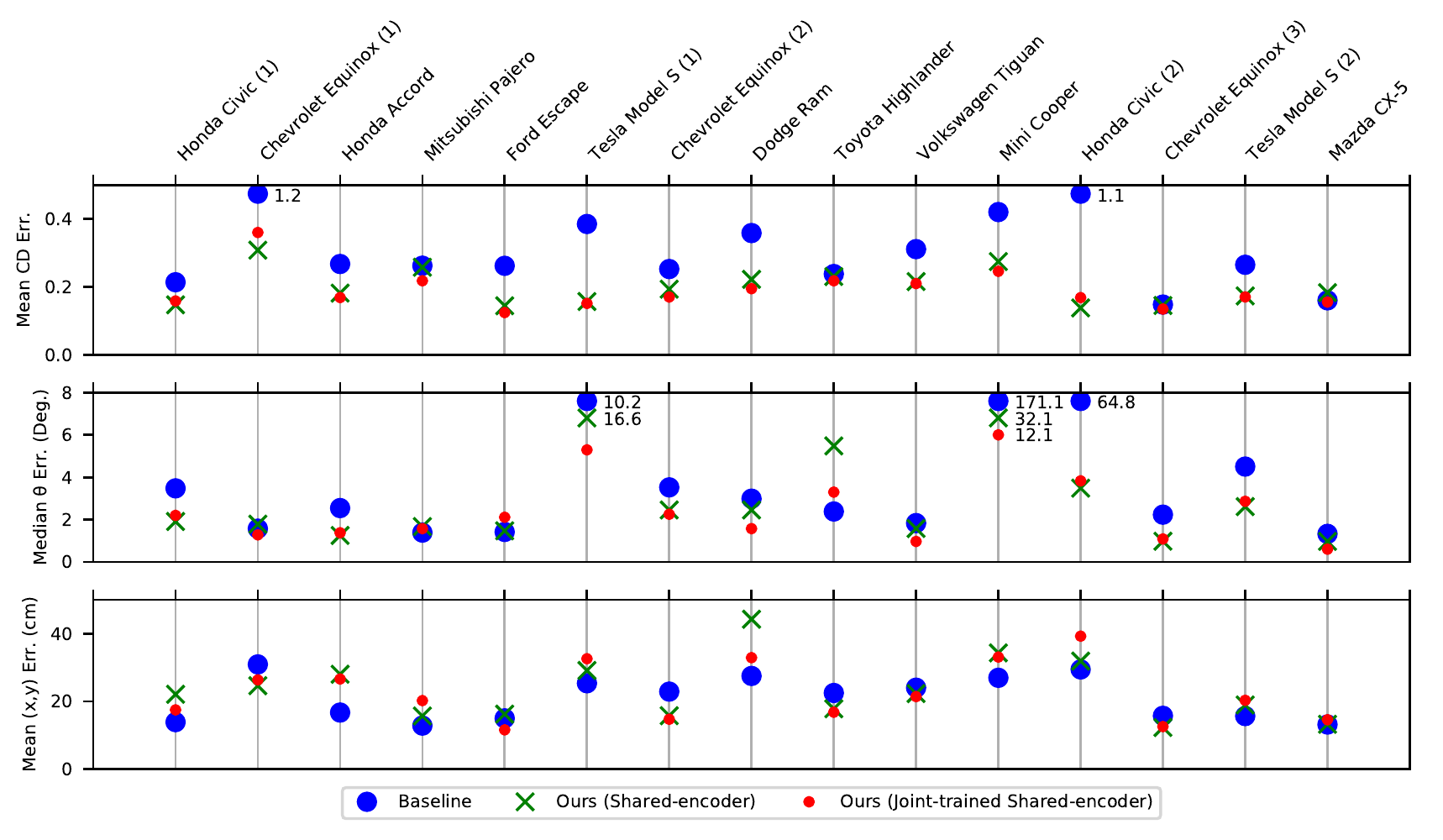}
    \caption{\textbf{Completion and pose accuracy on real-world Argoverse data.} Our shared-encoder and jointly-trained shared-encoder networks achieve good generalization performance on real-world data. We achieve lowest completion error for 12 out of 15 examples. The shared-encoder strategy for learning pose proves to be as good or better in many cases at estimating pose compared to the baseline. In total, we evaluate across 1,026 frames.
    }
    \label{fig:argoverse_statistics}
\end{figure*}

\subsection{Results}

Fig.~\ref{fig:hum3d_statistics} shows accuracy results for completion (CD) error and pose (heading, translation) error on our synthetic dataset. The results illustrate the performance improvement of the shared-encoder architecture over using separately-trained completion and pose networks as in the baseline network. Additionally, we find that the jointly-trained shared-encoder network has the best performance for both shape and pose estimation. We find the estimation of heading angle $\theta$ to be especially improved through our shared-encoder method, where information is shared between both pose and shape estimation tasks.

Fig.~\ref{fig:hum3d_qualitative} shows qualitative examples of pose and shape estimation on the validation set from our synthetic dataset. We provide examples showing performance across vehicle types, including rarely seen models. It it worth noting that we have trained a single network which performs shape and pose estimation for all examples shown.

\begin{figure}
    \centering
    \includegraphics[width=0.45\textwidth]{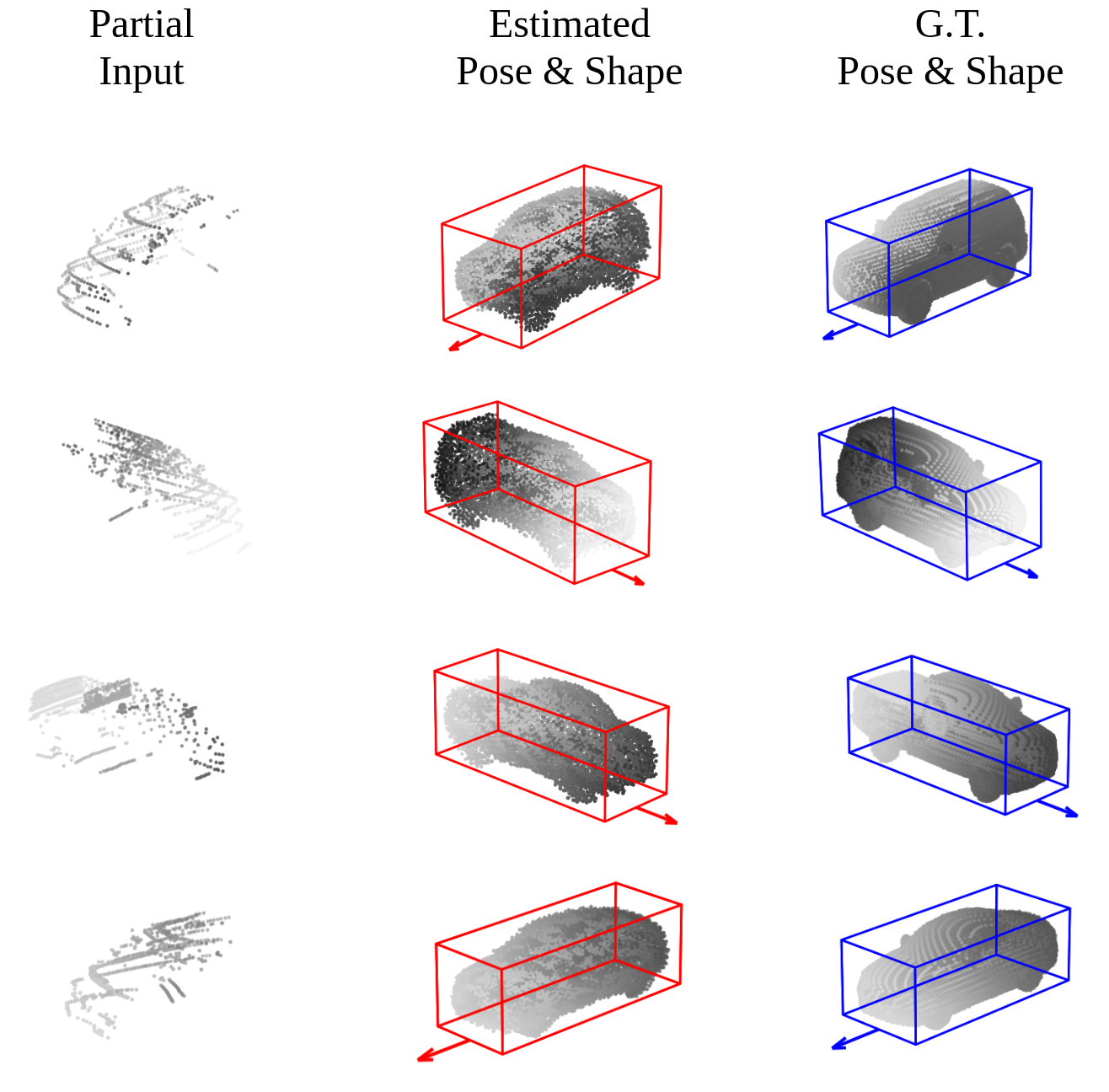}
    \caption{\textbf{Qualitative results on real-world Argoverse data.} Our joint-training method produces accurate pose and shape estimation on real-world data, despite being trained only on synthetic data. Shape estimation is accurate for different vehicle classes (SUV, pick-up truck, sedan). Heading angle $\theta$ estimation performs well without any cues from RGB data.\newline} 
    \label{fig:argoverse_qualitative}
\end{figure}

\section{Argoverse Experiment} \label{real_exp}

We compare performance of baseline, shared-encoder, and jointly-trained shared-encoder networks on real-world data from the recently released Argoverse dataset for tracking and motion forecasting~\cite{chang2019argoverse}. The Argoverse dataset is representative of real-world conditions for LiDAR capture, and provides a challenging test environment for the networks trained on synthetic data.

\subsection{Procedure}

We collected data from 15 different vehicle tracks in Argoverse for which we have access to the ground truth CAD model in our synthetic dataset, ensuring that the make, model, and year of the vehicles were closely matched to the available CAD models. We assume an ideal LiDAR segmentation algorithm by using the ground truth bounding box labels to segment the LiDAR points which are measured on the vehicle of interest. These unaligned LiDAR segmentations are used as input on which inference is done for each network. The resulting pose and shape estimations are then compared to the ground truth pose from the bounding box label and the ground truth complete point cloud, respectively. Across the 15 vehicle tracks, we evaluate a total of 1,026 frames.

\subsection{Results}

We provide results on the 15 different vehicle tracks in Fig.~\ref{fig:argoverse_statistics}, for both pose and shape estimation. Overall, we find impressive generalization of our shared-encoder and jointly-trained shared-encoder networks to real-world data, despite training on synthetic data. In addition, we find that the advantages over the baseline network which were seen in the synthetic case can generally be seen in the real data case as well. Specifically, we find that the shape estimates are closer to ground truth using our proposed models in 12 out of 15 tracks. Additionally, we find good performance pose estimation, despite sharing encoder parameters between the pose and shape estimation tasks. The results emphasize that the networks trained on synthetic data can be used effectively in real-world scenarios.

\begin{figure}
    \centering
    \includegraphics[width=0.45\textwidth]{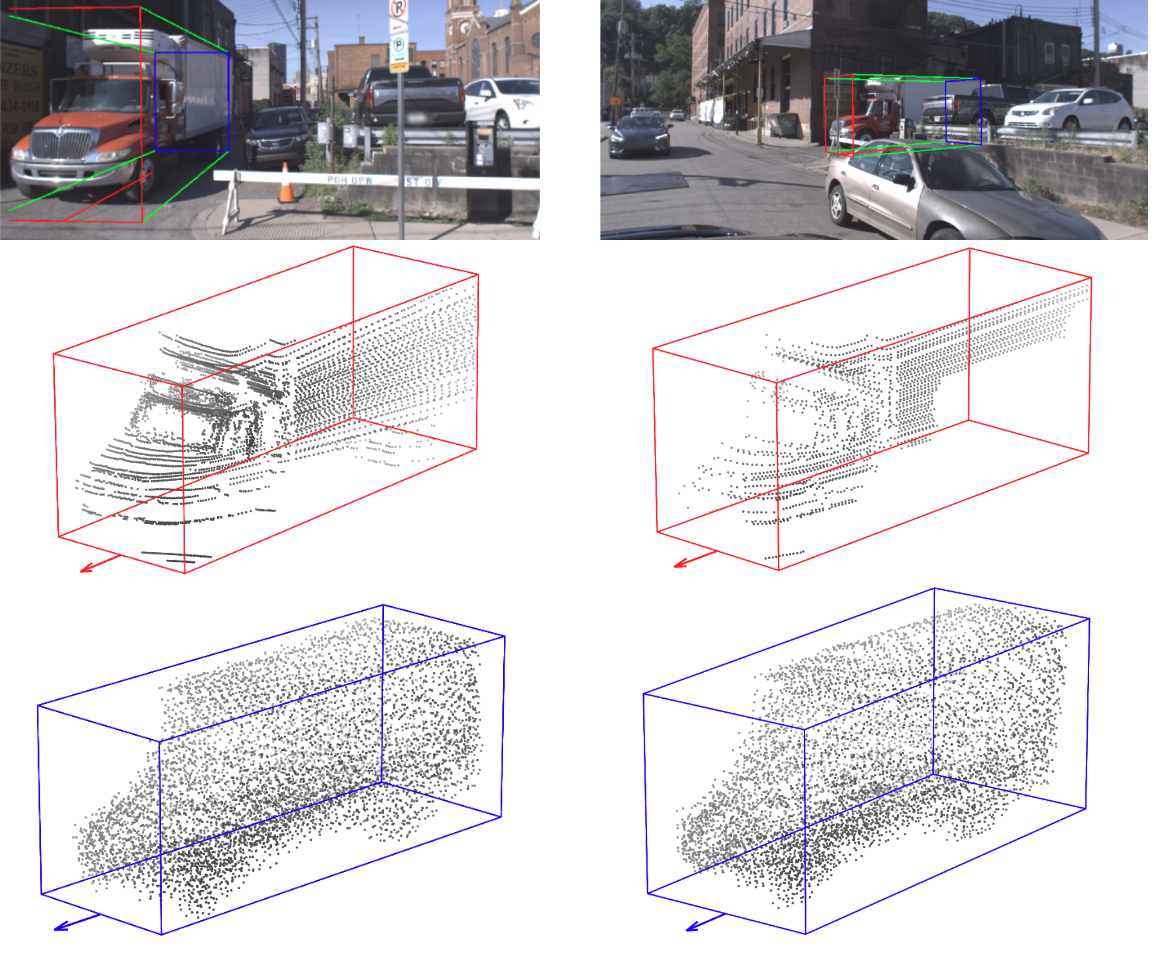}
    \caption{\textbf{Changing occlusion and point density.} We find that our shared-encoder architecture provides impressive consistency in pose and shape estimation on real-world data, despite drastic changes in occlusion and point density. This is illustrated by the differences in the LiDAR returns (middle row), while the estimated shape and pose (lowest row) remain consistent.}
    \label{fig:truck_qualitative}
\end{figure}

We further provide qualitative results from Argoverse in Fig.~\ref{fig:argoverse_qualitative} which illustrate shape and pose estimation performance on real-world data. Additionally, we find impressive robustness to occlusion, shown in Fig.~\ref{fig:truck_qualitative}. The shape and pose estimate for the semi-truck are consistent across large gaps in time, even with large occlusion and differing point densities. Additional qualitative videos may be found in the supplemental material.

\section{Discussion}

We have investigated a method of jointly determining shape and pose of vehicles from LiDAR data, a problem which is critical to autonomous vehicle applications but which has broader implications for many other problems. We have shown that information about pose and shape can effectively be shared and jointly-learned, to produce better performance in each task individually. We have adapted state-of-the-art deep-learning methods which use lightweight operations on point clouds, an approach which makes our method particularly effective in real-time systems.

We have not considered reflectance information of LiDAR returns, but this could plausibly offer cues as to the material property of vehicle surfaces, and therefore improve shape estimation. We have not explored non-vehicle agents which are commonly observed by autonomous vehicles, such as bicycles and pedestrians.

The proposed method has difficulty with lower numbers of LiDAR points, and the shape and pose estimation can degrade in such scenarios. A possible solution to this which could be further explored, is the integration of stereo camera data to offer additional 3D measurement of vehicles. With stereo input data, the use of RGB channel information may also be explored.

\section{Conclusion}

We have presented a joint method of pose and shape estimation for vehicles from LiDAR data, through which information is efficiently shared to simultaneously make predictions for each task. We believe that this work constitutes a valuable investigation in the joint learning of shape and pose recovery, and can constitute a valuable addition as a modular component to larger perception pipelines.

{\small
\bibliographystyle{ieee_fullname}
\bibliography{bib}
}

\end{document}